\documentclass[journal]{IEEEtran}

\usepackage{amssymb}
\usepackage{hyperref}
\usepackage[utf8]{inputenc}
\usepackage{amsfonts}
\usepackage{amsmath}
\usepackage{amssymb}
\usepackage{bm}
\usepackage{bookmark}
\usepackage{booktabs}
\usepackage{color}
\usepackage{commath}
\usepackage{graphicx}
\usepackage{mathtools}
\usepackage{makecell}
\usepackage{multirow}
\usepackage{subfig}
\usepackage{tabularx}

\DeclarePairedDelimiter\floor{\lfloor}{\rfloor}

\begin{document}
%
\title{A Survey of Dimensionality Reduction Techniques Based on Random Projection}


\author{\IEEEauthorblockN{Haozhe Xie, Jie Li, Hanqing Xue}
\thanks{H. Xie, J. Lie, and H. Xue are with  the School of Computer Science and Technology, Harbin Institute of Technology, Harbin 150001, China (email: hzxie@hit.edu.cn; jieli@hit.edu.cn; hqxue@stu.hit.edu.cn).}
}

%



\IEEEtitleabstractindextext{%
\begin{abstract}
Dimensionality reduction techniques play important roles in the analysis of big data.
Traditional dimensionality reduction approaches, such as principle component analysis (PCA) and linear discriminant analysis (LDA), have been studied extensively in the past few decades.
However, as the dimensionality of data increases, the computational cost of traditional dimensionality reduction methods grows exponentially, and the computation becomes prohibitively intractable. 
These drawbacks have triggered the development of random projection (RP) techniques, which map high-dimensional data onto a low-dimensional subspace with extremely reduced time cost. 
However, the RP transformation matrix is generated without considering the intrinsic structure of the original data and usually leads to relatively high distortion. 
Therefore, in recent years, methods based on RP have been proposed to address this problem. 
In this paper, we summarize the methods used in different situations to help practitioners to employ the proper techniques for their specific applications. 
Meanwhile, we enumerate the benefits and limitations of the various methods and provide further references for researchers to develop novel RP-based approaches.
\end{abstract}

}

\maketitle

\IEEEdisplaynontitleabstractindextext

%
\IEEEpeerreviewmaketitle

\section{Introduction}
%
%
%
%
\IEEEPARstart{T}{he} data in machine learning and data mining scenarios usually have very high dimensionality \cite{han2011data}. 
For example, market basket data in the hypermarket are high-dimensional, consisting of several thousand types of merchandise.
In text mining, each document is usually represented by a vector whose dimensionality is equals to the vocabulary size.
In bioinformatics, gene expression profiles can also be considered as matrices with more than ten thousand continuous values.
High dimensionality leads to burdensome computation and {\textit{curse-of-dimensionality}} issues.
Therefore, dimensionality reduction techniques are often applied in machine learning tasks to alleviate these problems \cite{roweis2000nonlinear}.
Traditional dimensionality reduction techniques, such as PCA \cite{wold1987principal} and LDA \cite{mika1999fisher}, have been widely studied in past decades.
However, as data dimensionality increases, the computational cost of traditional dimensionality reduction approaches grows exponentially, and the computation becomes prohibitively intractable.
RP \cite{dasgupta2000experiments}, which projects the original high-dimensional matrix $\mathbf{X}_{n \times d}$ onto a $k$-dimensional subspace using a random matrix $\mathbf{W}_{d \times k}$, is a simple and rapid approach to reduce dimensionality.
RP can be formulated as follows

\begin{equation}
	\mathbf{X}_{n \times k}^{RP} = \mathbf{X}_{n \times d}\mathbf{W}_{d \times k}
\end{equation}
The essential idea of RP is based on the Johnson-Lindenstrauss lemma \cite{johnson1984extensions}, which states that it is possible to project $n$ points in a space of arbitrarily high dimensions onto an $O(\log{n})$-dimensional space such that the pairwise distances between points are approximately preserved.
Thus, RP has attracted increasing attention in recent years and has been employed in many machine learning scenarios, including classification \cite{bingham2001random,xie2016comparison,zhao2015semi}, clustering \cite{fern2003random,sakai2009fast,tasoulis2014random},  and regression \cite{wan2014r3p,sun2015sprem,schneider2016forecasting}.
Although RP is much less expensive in terms of computational cost, it often fails to capture the task-related information because the latent space is generated without considering the intrinsic structure of the original data.
Various methods have been proposed to overcome this issue and to improve the performance of RP.
These methods can be classified into three categories: feature extraction approaches, dimensionality increasing approaches, and ensemble approaches.
Table~\ref{tab:approaches-to-improve-the-RP} provides a taxonomy of the approaches developed to improve the performance of RP, their most prominent advantages and disadvantages, and the corresponding literatures.

\begin{table*}[!htb]
    \caption{A taxonomy of approaches to improve the performance of RP. The advantages and disadvantages are listed along with the corresponding references.}
    \centering
    \begin{tabularx}{\linewidth}{lXXl}
    \toprule
    Approach                            & Advantages                    & Disadvantages                    & Ref. \\
    \midrule
    \multirow{4}{*}{Feature extraction} & \multicolumn{3}{l}{General-purpose methods} \\
                                          \cmidrule(lr){2-4}
                                        & \makecell[tl]{Applicable to most datasets \\ Good at finding discriminative features }
                                        & \makecell[tl]{Computationally intensive}
                                        & \makecell[tl]{\cite{xie2016comparison} \\ \cite{zhao2015semi} \\ \cite{schneider2016forecasting} \\ \cite{liu2012sorted} \\ \cite{wu2016random}} \\
                                        & \multicolumn{3}{l}{Application-specific methods} \\
                                          \cmidrule(lr){2-4}
                                        & \makecell[tl]{Better at finding discriminative features than \\ general-propose methods}
                                        & \makecell[tl]{Computationally intensive \\ Applicable to a few datasets}
                                        & \makecell[tl]{\cite{arriaga2015visual} \\ \cite{qin2015multiscale} \\ \cite{feng2015random} \\ \cite{du2011random} \\ \cite{majee2015efficient}} \\
    \midrule
    Dimensionality increasing           & \makecell[tl]{Fast \\ Improves linear separability \\ }
                                        & \makecell[tl]{Bad at fitting complex features \\ Weak in finding discriminative features}
                                        & \makecell[tl]{\cite{ma2015random} \\ \cite{alshamiri2015combining} \\ \cite{shan2016improved} \\  \cite{zhang2012real} \\ \cite{zhang2014fast}} \\
    \midrule
    Ensemble                            & \makecell[tl]{Robust \\ Lower risk of overfitting \\ Applicable to most datasets \\ Performs well on imbalanced datasets}
                                        & \makecell[tl]{Computationally intensive \\ Slow in making predictions \\ Sensitive to noise (Boosting)}
                                        & \makecell[tl]{\cite{schclar2009random} \\ \cite{zhang2017drugrpe} \\ \cite{yoshioka2013evaluation} \\ \cite{gondara2015rpc}} \\
    \bottomrule
    \end{tabularx}
    \label{tab:approaches-to-improve-the-RP}
\end{table*}

Feature extraction approaches, which are the most commonly used approach to improve the performance of RP, attempt to construct informative and non-redundant features from a large set of data.
These methods can be divided into two major categories: general-propose methods and application-specific methods.
Generally, application-specific feature extraction methods find better discriminative features than general-propose methods do, but they are limited to a small number of datasets.
The main drawback of feature extraction approaches is that they are usually computationally intensive.

Dimensionality increasing approaches map a low-dimensional feature space onto a higher-dimensional feature space while improving the linear separability. 
The original features can be better represented in high-dimensional space \cite{lu2017kernel}.
The generated high-dimensional space requires impracticably large computational resources, so RP is used for dimensionality reduction.
According to the existing literatures on RP, extreme learning machine (ELM) \cite{huang2006extreme} and rectangle filters \cite{viola2001rapid} are often used to increase the dimensionality of the original feature space.
Both methods are computationally fast; however, their architecture is so simple that they often have trouble fitting complex features \cite{cao2015extreme}.

Ensemble approaches have been studied extensively, including the well-known random forest \cite{liaw2002classification} and AdaBoost \cite{freund1999short}. 
These methods are robust and perform well on imbalanced datasets \cite{galar2012review}.
Also, ensembles of multiple RP instances lead to lower risks of overfitting and better generalization performance.
Nevertheless, as the data dimensionality increases, prediction becomes incredibly slow.

Although various approaches have been developed to improve the performance of RP, some issues still need to be addressed. 
Practitioners also require guidelines to select the proper approach to use for their specific application. 
Here, we review these approaches and summarize their benefits and limitations to provide a reference for further studies of RP-based methods.

\section{Feature extraction approaches}

\begin{figure*}[!htb]
  \centering
  \resizebox{\linewidth}{!} {
    \includegraphics{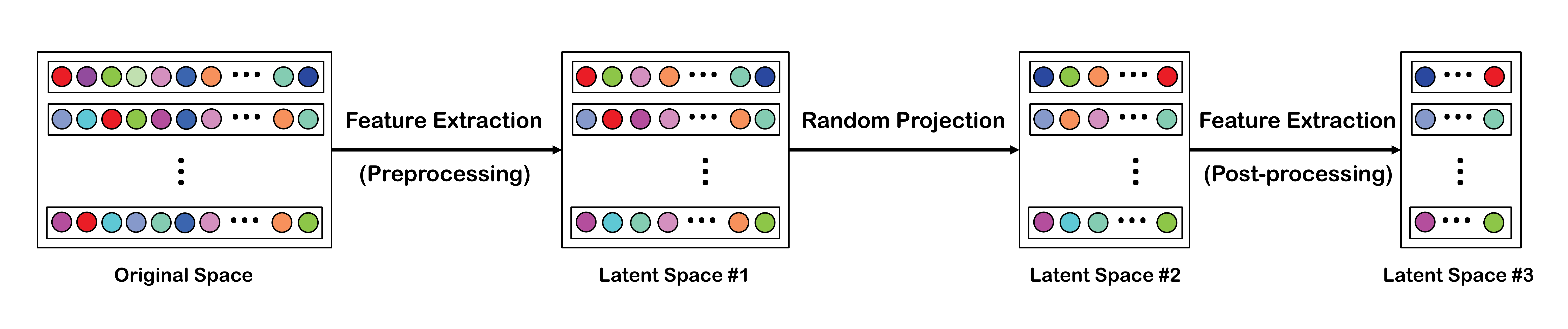}
  }
  \caption{Preprocessing and post-processing methods for RP. Both of them are used for feature extraction and help RP to better capturing the intrinsic structure of the original data.}
  \label{fig:feature-extraction-methods-overview}
\end{figure*}

\begin{table*}[ht]
    \caption{Alternative preprocessing and post-processing feature extraction methods used to improve the performance of RP.}
    \centering
    \begin{tabularx}{\linewidth}{XXXl}
    \toprule
    Preprocessing         & Post-processing       & Application                    & Ref. \\
    \midrule
    PCA; FS; (None)       & PCA; LDA; FS; (None)  & Tumor tissue classification    & \cite{xie2016comparison} \\
    (None)                & LDA                   & Text classification            & \cite{zhao2015semi} \\
    BoW                   & (None)                & Sales rank prediction          & \cite{schneider2016forecasting} \\
    (None)                & BoW                   & Texture image classification   & \cite{liu2012sorted} \\
    CNN                   & RNN                   & Text recognition in images     & \cite{wu2016random} \\
    FAST Corder Detector  & (None)                & Image classification           & \cite{arriaga2015visual} \\
    NSP                   & Mahalanobis distance  & Target enhancing               & \cite{qin2015multiscale} \\
    (None)                & CEM                   & Hyperspectral target detection & \cite{feng2015random} \\
    (None)                & TCIMF                 & Hyperspectral target detection & \cite{du2011random} \\
    (None)                & MUSIC                 & Wideband spectrum sensing      & \cite{majee2015efficient} \\
    \bottomrule
    \end{tabularx}
    \label{tab:feature-extraction-methods-summary}
\end{table*}

Feature extraction transforms data from high-dimensional space lower-dimensional space, which is the most commonly used approach to improve the performance of RP.
It is often used before or after RP as a preprocessing or post-processing step, respectively (Figure~\ref{fig:feature-extraction-methods-overview}).
A general overview of preprocessing and post-processing methods for RP in different application fields is presented in Table \ref{tab:feature-extraction-methods-summary}.
Researchers and practitioners prefer to use feature extraction in the post-processing stage because RP reduces the dimensionality of the feature space, which greatly accelerates the feature extraction methods.
In many machine learning and pattern recognition systems, feature extractors that transform raw data into feature vectors must be carefully designed, especially those in computer vision, including histogram of oriented gradient (HOG) \cite{dalal2005histograms} and scale invariant feature transform (SIFT) \cite{lowe2004distinctive}.
So important are Feature extractors that they directly affect the performance of the developed methods.
According to extant literatures on RP, the use of feature extraction methods can be roughly divided into two categories: 
general-purpose methods, such as some statistical learning methods and neural networks, which can be applied to many fields (e.g., natural language processing and computer vision); and application-specific methods, which are designed to address a specific problem, like nonsubsampled pyramid (NSP) \cite{da2006nonsubsampled} and constrained energy minimization (CEM) \cite{harsanyi1993detection}.

\subsection{General-purpose methods}
\label{sec:general-purpose-methods}

Traditional dimensionality reduction techniques search for dimensions with the maximum discriminative power, whereas RP performs well in rapidly finding low-dimensional space.
It is natural to combine the two types of methods to solve dimensionality reduction problems.
Thus, in the past few years, substantial research based on the two techniques has been conducted.

Xie et al. \cite{xie2016comparison} incorporated RP into PCA, LDA, and feature selection (FS) \cite{student1908probable} to classify gene expression profiles of breast cancer.
The experimental results demonstrated that the classification accuracy of RP can be significantly improved by FS, especially in small-n-large-p datasets \cite{antoniadis2003effective}.

\begin{figure*}[!htb]
  \centering
  \resizebox{.75\linewidth}{!} {
    \includegraphics{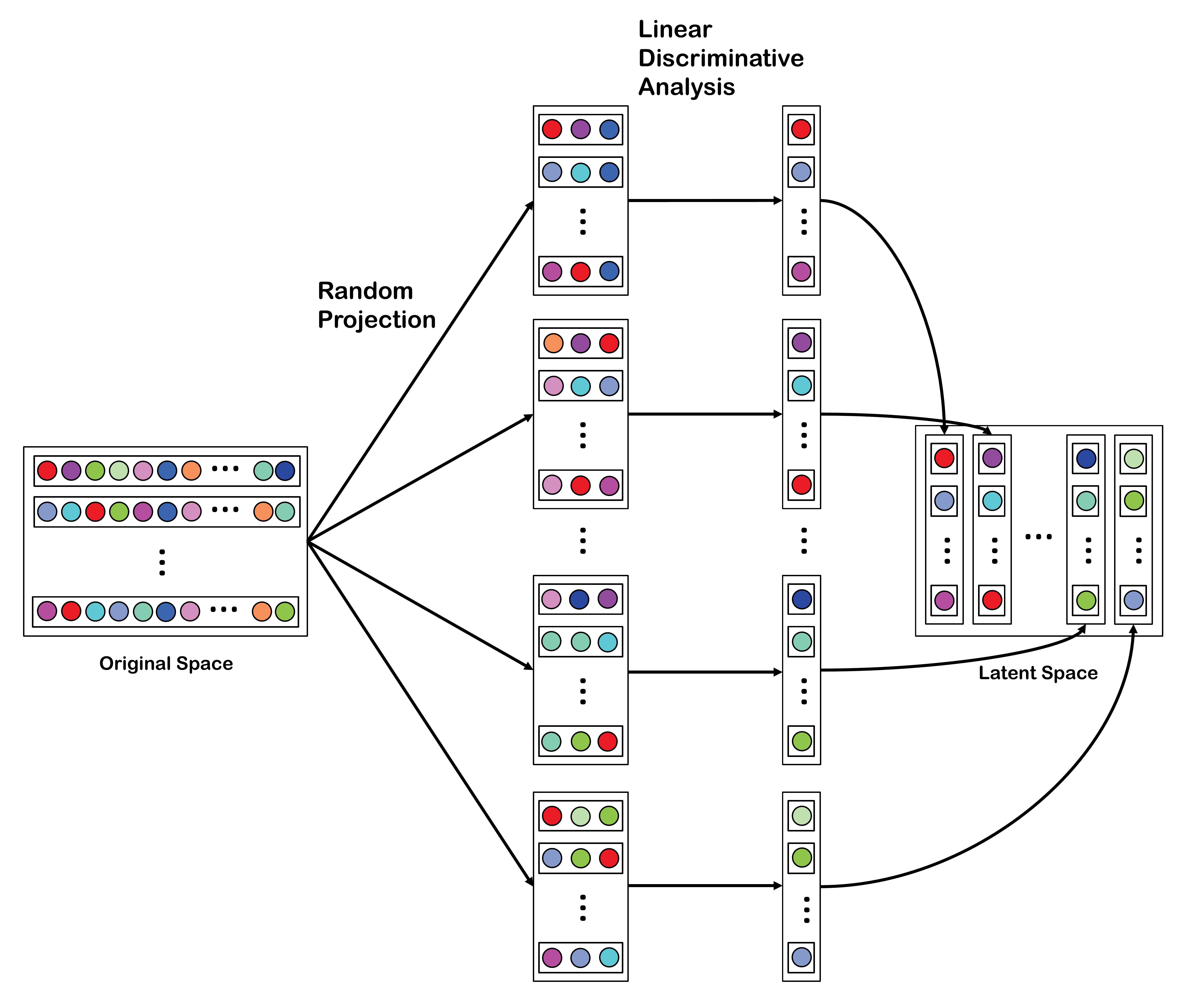}
  }
  \caption{Unified graph explanation for random projection and semi-random projection. The latent space consists of features generated by different RP transformation matrices.}
  \label{fig:semi-random-projection-overview}
\end{figure*}

Zhao et al. \cite{zhao2015semi} proposed semi-random projection (SRP)  to find a discriminative subspace while maintaining a feasible computational load.
In contrast to RP, where the values of the transformation matrix are assigned randomly, the weights for the transformation vectors of SRP are obtained by LDA.
The SRP method (see Figure~\ref{fig:semi-random-projection-overview}) consists of three steps.
First, the original data matrix $\mathbf{X} \in \mathbb{R}^{n \times d}$ is mapped onto a subspace $\widehat{\mathbf{X}_i} \in \mathbb{R}^{n \times k}$ using the randomly selected $k$ features.
Next, the data with $k$ features are projected onto a single dimension $h_i$ using a transform vector $\mathbf{W} \in \mathbb{R}^{k \times 1}$ learned by LDA.
The above procedure is repeated $r$ times to generate the following latent subspace $\mathbf{H} \in \mathbb{R}^{n \times r}$

\begin{equation}
	\mathbf{H} = \begin{bmatrix} \mathbf{h}_1 & \mathbf{h}_2 & \dots & \mathbf{h}_r \end{bmatrix}
\end{equation}
Experiments were performed on six datasets generated from the standard text 20 newsgroups corpus \cite{lang1995newsweeder} for determining the categories of texts.
The experimental results indicated that the classification accuracy of SRP followed by PCA (SRP + PCA) increases by approximately 25\% compared to that of RP, but it is still lower than that of PCA.

\begin{figure*}[!htb]
  \centering
  \resizebox{.5\linewidth}{!} {
    \includegraphics{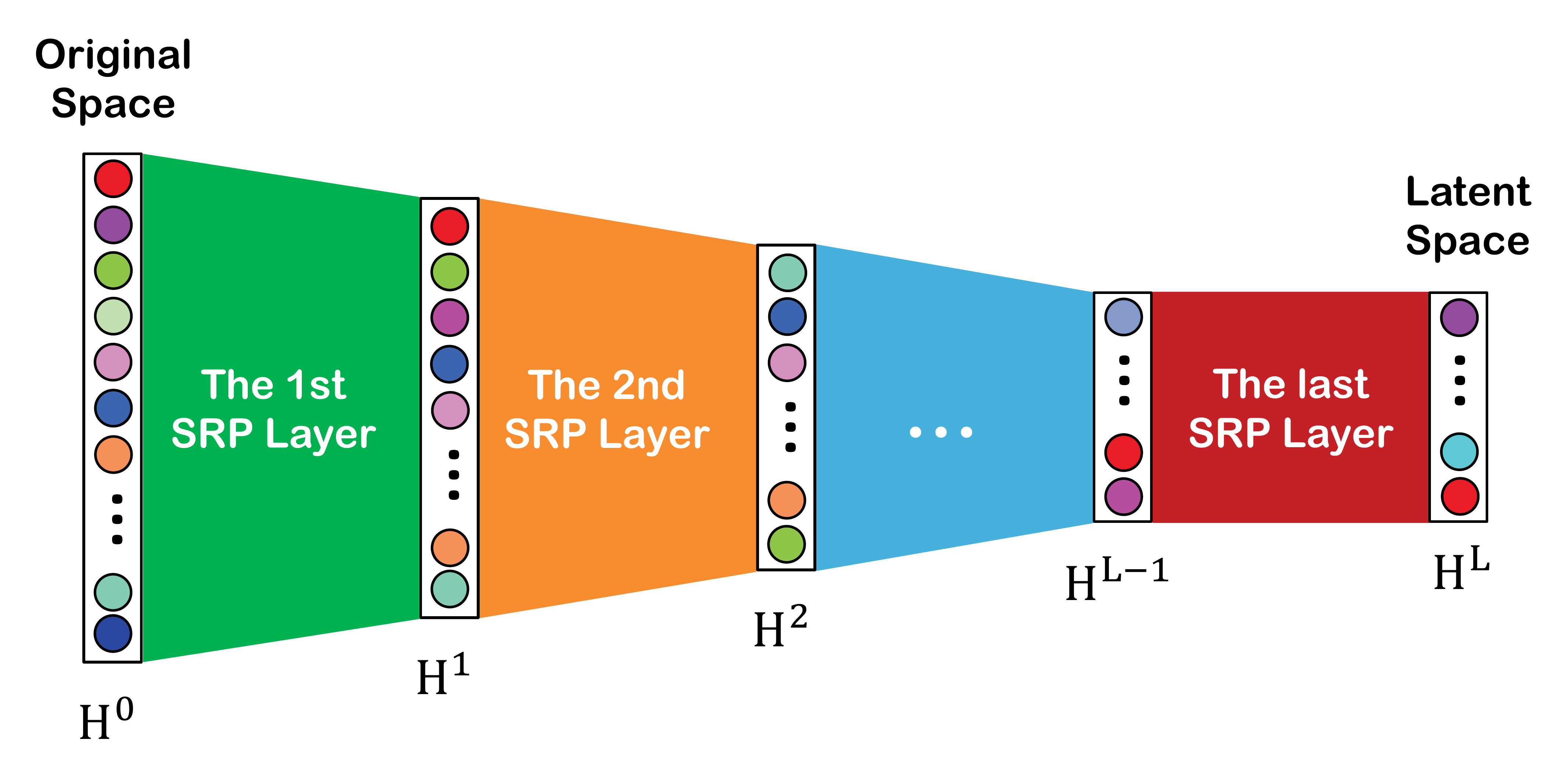}
  }
  \caption{Architecture of stacked semi-random projection with $L$ layers, in which SRP is stacked layer by layer. The dimension in the left several layers becomes smaller and smaller. The last several layers keep the dimension unchanged.}
  \label{fig:stacked-semi-random-projection-overview}
\end{figure*}

Stacked semi-random projection (SSRP) was developed to further improve the performance of SRP.
In SSRP, SRP is stacked by feeding the output of the previous SRP layer as the input of the subsequent SRP layer.
Suppose the output of the $k$-th SRP layer is denoted by $\mathbf{H}^k$ with dimensionality of $d^k$ and that the first layer is initialized with the original data $\mathbf{H}^0 = \mathbf{X}$ with dimensionality of $d^0$.
The relationship between the dimensionality of the $k$-th and $(k-1)$-th SRP layer can be calculated as

\begin{equation}
    d^k = \alpha \floor{\sqrt{d^{k - 1}}}
\end{equation}
where $\alpha$ is a hyperparameter.
The output of the $k$-th SRP layer can be determined as

\begin{equation}
	\mathbf{H}^k = \frac{1}{1 + \exp (-[\mathbf{W}^k]^{\mathrm T} \mathbf{H}^{k - 1})}
\end{equation}
where $\mathbf{W}^k$ is the transformation vector of the k-th SRP layer learned from regularized LDA \cite{friedman1989regularized}.
The overall process of SSRP is shown in Figure~\ref{fig:stacked-semi-random-projection-overview}.
SSRP outperforms other methods, including PCA, sparse PCA \cite{d2007direct}, RP, RP + PCA, mSDA \cite{chen2012marginalized}, and SRP + PCA, on 5 of 6 datasets in terms of classification accuracy.
In addition, the computational cost of SSRP is much lower than that of PCA, although it is 2 times higher than that of RP.
Because LDA is adopted, SRP and SSRP may easily overfit the data in the presence of labeling noise and are not applicable for non-linear problems \cite{yan2011comparison}.

The BoW model is a simplified representation used in natural language processing and computer vision \cite{liu2012sorted,alavi2014random,maqueda2015novel}.
The number of words in a BoW model usually reaches tens of millions, which explodes the dimensionality of predictor matrices.
In the past few years, RP has been used to address this issue.
Schneider et al. \cite{schneider2016forecasting} proposed an attributes-based regression model in which historical data were used to forecast one-week-ahead rolling sales for tablet computers.
In contrast to the extant approaches \cite{chevalier2003measuring,archak2011deriving}, the authors considered customer reviews and used a BoW model to analyze product feedback.
However, a key challenge of BoW is that the millions of words contained in the bag lead to infeasible computation.
RP was adopted to reduce the dimensionality of the BOW predictor matrices to address this problem.
Assume that the BoW matrix is denoted by $\mathbf{BOWS}_{n \times d}$, where $n$ and $d$ are the number of reviews and words, respectively.
RP maps the original space onto a subspace $\widehat{\mathbf{BOWS}}_{n \times k}$ spanned by the $k$ features.
Compared to the baseline model without BoW, the proposed model with BoW \cite{alavi2014random} has better predictive performance on the dataset named ``Market Dynamics and User Generated Content about Tablet Computers'' \cite{wang2013database}. 
Since the BoW representation ignores the semantic relation between words, it may fail to capture order information and dependency between the words in reviews \cite{kosala2000web}.

In computer vision, the BoW model has been applied to texture research (including synthesis, classification, segmentation, compression, and shape from texture) \cite{leung2001representing,varma2005statistical,zhang2007local,varma2009statistical} by treating texture features as words.
In BoW, texture images are statistically described as histograms over a dictionary of features.
It has been proved that local texture feature descriptors learned from BoW are insensitive to local image perturbations such as rotation, affine changes and scale \cite{liu2012sorted}.
Liu et al. \cite{liu2012texture} proposed a robust and powerful texture classifier based on RP and BoW.
First, RP is used to extract a small set of random features from local image patches.
Then, random features are embedded into a BoW model to conduct texture classification.
Finally, learning and classification are performed in the compressed domain.
The proposed method improves the classification accuracy by 10.38\% and 3.32\% compared to local binary pattern (LBP) \cite{ojala2002multiresolution} and the combination of LBP and normalized Gabor filter (NGF) \cite{clausi2005design}.
Furthermore, the proposed method consumes considerably less time and storage space. 
However, BoW may still reduce the discriminative power of images because it ignores the geometric relationships among visual words \cite{tsai2012bag}.

Traditional machine learning techniques are limited in their ability to handle natural data in their raw form.
Deep learning \cite{hinton2006fast} has performed remarkably well, leaving traditional machine learning in the dust.
Wu et al. \cite{wu2016random} proposed a novel method based on RP and deep neural network (DNN) \cite{krizhevsky2012imagenet} for text recognition in natural scene images.
First, a convolutional neural network (CNN) \cite{krizhevsky2012imagenet} is adopted as a feature extractor to convert word images into a multi-layer CNN feature sequence with slicing windows.
Then, RP is used to map the high-dimensional CNN features onto a subspace with low dimensionality.
Finally, a recurrent neural network (RNN) \cite{mikolov2010recurrent}, used for decoding the embedded RP-CNN features, is trained to recognize the text in the image.
Multiple RNNs are ensembled by the Recognizer Output Voting Error Reduction (ROVER) algorithm \cite{fiscus1997post} to further improve the recognition rate.
The authors found that the recognition rate of RP-CNN features, with 85\% dimensionality reduction, is similar to that of the original high-dimensional features.
Although CNNs and RNNs dramatically improves the recognition rate, CNNs and RNNs are difficult to train and require substantial amounts of time to select the hyperparameters \cite{lecun2015deep}.

Low-rank approximation plays a central role in data analysis.
In mathematics, it is often desirable to find a good approximation of a given matrix with a lower rank.
Eigenvalue decomposition is a typical strategy, for example, the best known method is singular value decomposition (SVD) \cite{golub1970singular}, but it usually leads to heavy computation.
RP is a simple technique and has been widely used to accelerate the computations for such approximations.
Assume that the original matrix can be formulated as $\mathbf{X} \in \mathbb{R}^{n \times k}$.
The quality of the approximation depends on how well the method captures the important part of $\mathbf{X}$.
Martinsson \cite{martinsson2016randomized} proposed an approximation method based on QR decomposition \cite{horn2012matrix} and RP.
First, a matrix $\mathbf{Y} \in \mathbb{R}^{n \times d}$ is formed according to $\mathbf{Y}=(\mathbf{X}\mathbf{X}^{\mathrm T})^p\mathbf{X}\mathbf{S}$, where $p$ stands for the number of iterations, and  $\mathbf{S} \in \mathbb{R}^{k \times d}$ follows Gaussian distribution $\mathcal{N}(0,1)$.
Then, QR decomposition is applied: $\mathbf{Y} = \mathbf{Q}\mathbf{R}$.
The low-rank approximation of matrix $\mathbf{X}$ can be represented as $\widetilde{\mathbf{X}} = \mathbf{Q}(\mathbf{Q}^{\mathrm T}\mathbf{X})$.
The proposed method constructs a nearly optimal rank-k approximation with much lower time complexity.
Alternatively, the sampling matrix $\mathbf{S}$ can be sampled from subsampled randomized Hadamard transform (SRHT) \cite{tropp2011improved}, and the corresponding approximation method is known as structured RP \cite{woolfe2008fast}.
The experimental results revealed that the structured RP is faster but less accurate than the method proposed by Martinsson \cite{martinsson2016randomized}.

Zhang et al. \cite{zhang2016accelerated} proposed a method to accelerate linear regression with the help of  low-rank matrix approximation implemented with RP and QR decomposition.
Suppose the observed data matrix and the corresponding response vector are represented as $\mathbf{X}$ and $\mathbf{Y}$, respectively.
Let $\widetilde{\mathbf{X}}$ be the low-rank approximation of $\mathbf{X}$; thus, the coefficient vector ${\bm \beta}$ can be calculated according to the following recursive formula

\begin{equation}
\begin{aligned}
    {\bm \beta}_{t + 1} = \operatorname*{argmin}\limits_{{\bm \beta} \in \mathbb{R}^d} \Biggl(-\frac{1}{n} ({\bm \beta} - {\bm \beta}_t)^{\mathrm T} \widetilde{\mathbf{X}} (\mathbf{Y} - \widetilde{\mathbf{X}}^{\mathrm T}{\bm \beta}_t) \\
    + \Phi_t \left\Vert{\bm \beta}\right\Vert_1 + \frac{\lambda}{2} \left\Vert{\bm \beta} - {\bm \beta}_t\right\Vert_2^2\Biggr)
\end{aligned}
\end{equation}
where $\lambda$ represents the regularization term.
${\bm \beta}_0$ is set to $\mathbf{0}$, and $\Phi_t= \min(\Phi_{min},\Phi_0\eta_t)$, in which $\eta_t \in (0, 1)$ controls the shrinkage speed of $\Phi_t$.
Compared to existing methods excluding RP, such as PGH \cite{xiao2013proximal} and ADG \cite{nesterov2013gradient}, the proposed method significantly reduces the computation time by 98.90\% and 98.55\%.
Additionally, the convergence rate of the proposed method is comparable to that of PGH and much higher than that of ADG.

\subsection{Application-specific methods}

\begin{figure*}[!htb]
  \centering
  \resizebox{.85\linewidth}{!} {
    \subfloat[Sliding window random projection]{
      \includegraphics{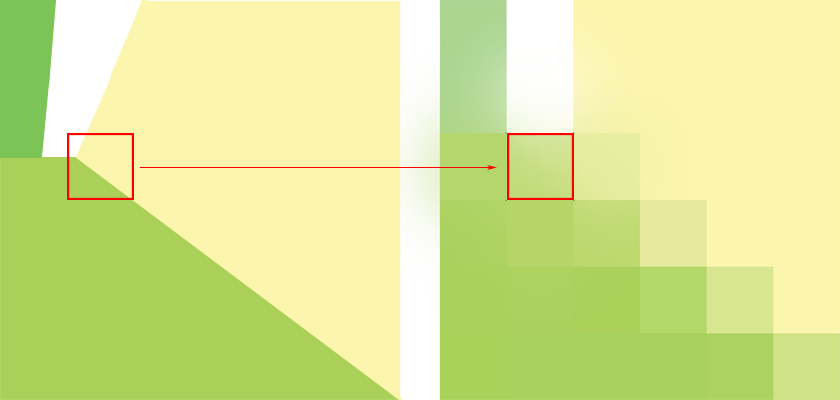}
      \label{fig:slide-window-random-projection}
    }
    \subfloat[Corner random projection]{
      \includegraphics{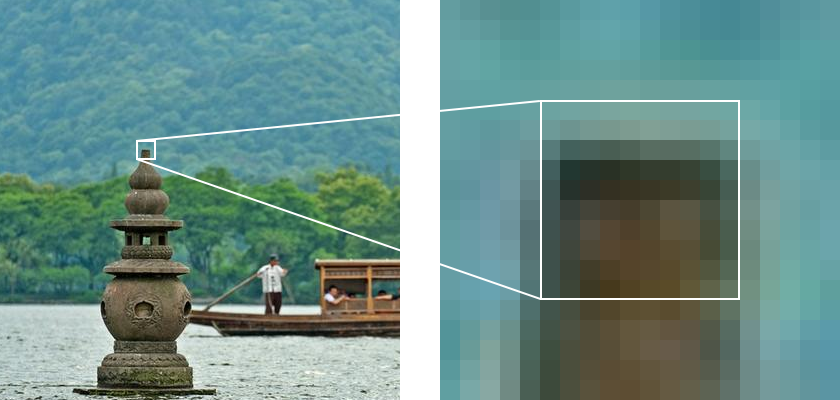}
      \label{fig:corner-random-projection}
    }
  }
  \caption{Description of (a) sliding window random projection and (b) corner random projection. In sliding window random projection, features are represented by a random combination of colors within a sliding window. In corner random projection, the features are represented by corners of objects.}
\end{figure*}

In some specific application fields, especially image processing, specific feature extraction methods are often designed to obtain better experimental results.
Arriaga et al. \cite{arriaga2015visual} proposed two RP-based methods: sliding window RP (SWRP) and corner RP (CRP).
In SWRP, a corresponding location in the projected images is filled with a color that generated from a random combination of colors in a window that slides over the original image (see Figure~\ref{fig:slide-window-random-projection}).
Motivated by the intuition that human visual systems are feature detectors for colors, lines, corners, and so on, the features from accelerated segment test (FAST) \cite{rosten2006machine} corner detector is adopted in CRP to detect features in images.
The projected image can be represented as a combination of multiple corner images (see Figure~\ref{fig:corner-random-projection}).
The authors indicated that SWRP performs better than CRP in terms of classification accuracy; 
neural networks without SWRP have similar classification accuracy but much lower computational cost than neural networks with SWRP.
However, the features extracted from SWRP and CRP are not rotation-invariant because neither introduces multi-scale features.

\begin{figure*}[!htb]
  \centering
  \resizebox{.7\linewidth}{!} {
    \includegraphics{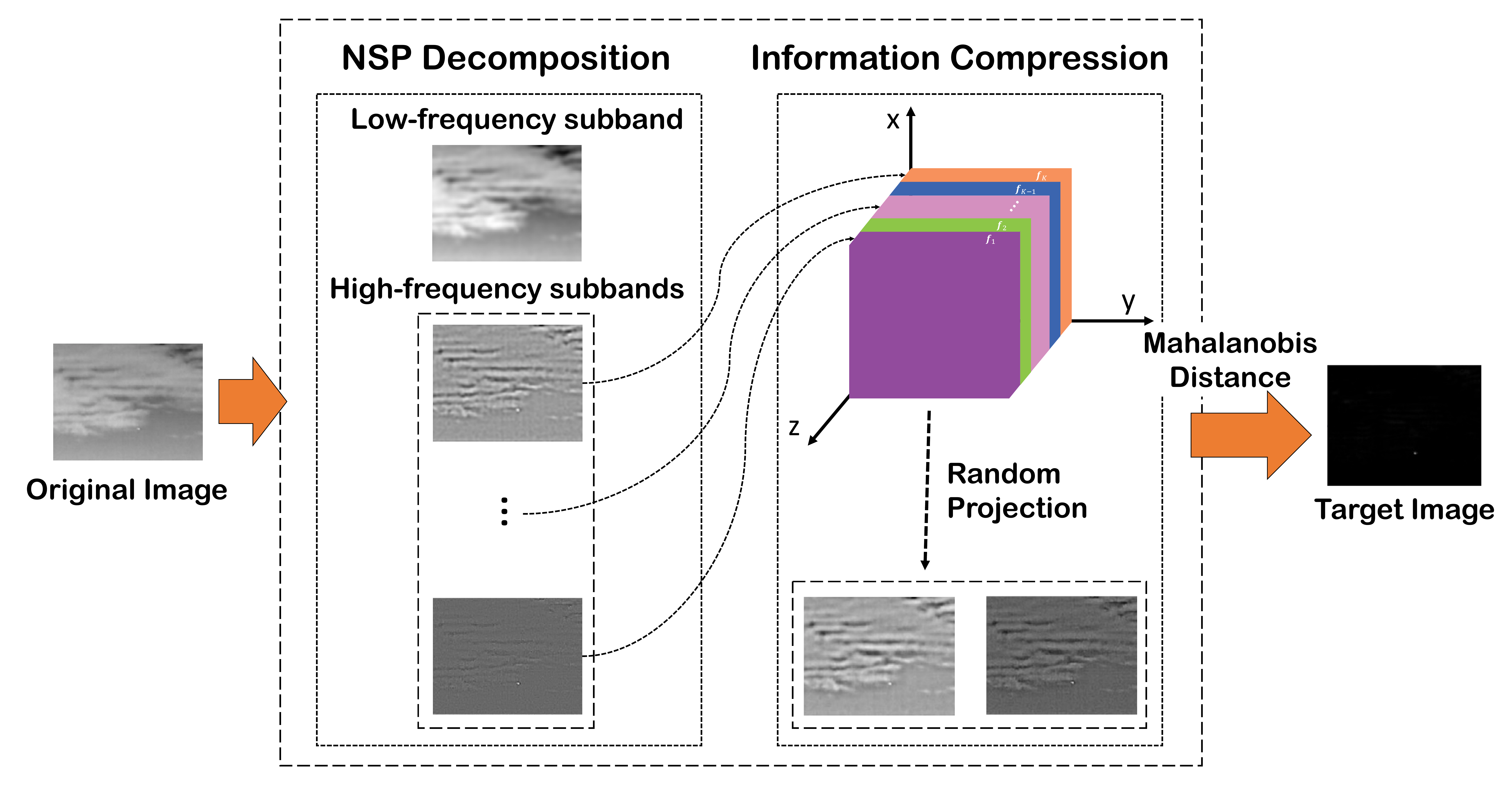}
  }
  \caption{Flowchart of the background suppression method, where NSP is used for feature extraction and RP is used for information compression.}
  \label{fig:background-suppression-method-overview}
\end{figure*}

By taking advantage of the fact that RP is insensitive to noise in images, Qin et al. \cite{qin2015multiscale} proposed a method to suppress clutters while enhancing targets in infrared images (see Figure~\ref{fig:background-suppression-method-overview}).
First, a signal decomposition algorithm named NSP \cite{da2006nonsubsampled} is adopted to decompose an image into a single low-frequency subband and multiple high-frequency subbands.
After $K$-scale NSP decomposition, $K + 1$ subbands, including one low-frequency subband and $K$ high-frequency subbands, are produced.
The high-frequency subbands mainly contain targets and minimally cluttered backgrounds, so it is easier to extract target information from the high-frequency subbands.
To preserve as much target information as possible, a 3D image cube is constructed by concatenating all the high-frequency subbands. The cube model can be expressed as

\begin{equation}
    \mathbf{F}_{K \times M} = (f_1, f_2, \dots, f_k, \dots, f_K)^{\mathrm T}
\end{equation}
where $f_k$ is the row vector representation of a subband and $M$ is the number of pixels of the $k$-th high-frequency subband.
Then, RP is used to project the 3D image cube $\mathbf{F}_{K \times M}$ onto a low-dimensional subspace $\mathbf{Q}_{S \times M}$ to reduce the spatial redundancy of the target and background information.
Finally, the Mahalanobis distance \cite{de2000mahalanobis} is applied to remove background clutter in dimensionality-reduced high-frequency subbands obtained above.
Compared to other state-of-the-art methods for the background suppression of infrared small-target images, the proposed model outperforms max-median \cite{deshpande1999max}, morphological (top-hat) \cite{tom1993morphology}, phase spectrum of quaternion Fourier transform (PQFT) \cite{guo2008spatio} and wavelet transform (WRX) \cite{daubechies1990wavelet} in terms of signal-to-clutter ratio gain (SCRG) and background suppression factor (BSF) \cite{yang2015directional}.

Feng et al. \cite{feng2015random} developed a new approach to detect hyperspectral targets using the CEM method \cite{harsanyi1993detection}.
The objective of CEM is to design a finite impulse response (FIR) linear filter with $L$ filter coefficients, and the FIR filter can be represented using an $L$-dimensional vector $\mathbf{w} = \left\{w_1, w_2, \dots, w_L\right\}$. Let $r_i  (i=1, 2, \dots, n)$ be an $L$-dimensional sample pixel vector.
The optimized target detector $\mathbf{w}$ can be given as

\begin{equation}
    \mathbf{w}^* = \frac{\mathbf{R}^{-1}_{L \times L} \mathbf{d}}{\mathbf{d}^{\mathrm T} \mathbf{R}^{-1}_{L \times L} \mathbf{d}}
    \label{eq-cem-detector}
\end{equation}
where $\mathbf{R}_{L \times L} = (1/n)\sum_{i=1}^n \mathbf{r}_i\mathbf{r}_i^{\mathrm T}$ is the sample autocorrelation matrix of the target and $\mathbf{d}$ denotes the spectral signature of the target.
RP is incorporated to resolve the issue of the {\textit{curse of dimensionality}} in hyperspectral imagery.
The noise suppression effect of RP is superior to that of the maximum-noise-fraction (MNF) \cite{green1988transformation} and PCA, and therefore the CEM method with dimensionality reduction by RP (RP-CEM) outperforms MNF-CEM and PCA-CEM in terms of both detection accuracy and computation time.
The drawback of CEM is that it can detect only a single target because CEM treats undesired targets as interference and does not make full use of the known information.
The target-constrained interference-minimized filter (TCIMF) \cite{ren2000target} was developed to address this issue.
It assumes the pixels of an image are composed of three separate signal sources, $\mathbf{D}$ (desired targets), $\mathbf{U}$ (undesired targets), and $\mathbf{I}$ (interference), whereas CEM simply treats $\mathbf{U}$ as a part of $\mathbf{I}$.
Let $\mathbf{D} = \left[d_1, d_2, \dots, d_p\right]$ and $\mathbf{U} = \left[u_1, u_2, \dots, u_q\right]$ denote the desired-target signature and undesired-target signature, respectively.
The spectral signature of the target $\mathbf{d}$ in Eq.~\ref{eq-cem-detector} can be replaced with $\left[\mathbf{D}, \mathbf{U}\right]$. The optimal weight vector $\mathbf{w}$ can be formulated as

\begin{equation}
    \mathbf{w}^* = \frac{\mathbf{R}^{-1}_{L \times L} \left[\mathbf{D}~\mathbf{U}\right]}{\left[\mathbf{D}~\mathbf{U}\right]^{\mathrm T} \mathbf{R}^{-1}_{L \times L} \left[\mathbf{D}~\mathbf{U}\right]} 
    \begin{bmatrix}
    1_{p \times 1} \\
    0_{q \times 1}
    \end{bmatrix}
\end{equation}
where $1_{p \times 1}$ is a $p \times 1$ column vector with ones in its components that is used to constrain the desired targets in $\mathbf{D}$.
Similarly, $0_{q \times 1}$ is a $q \times 1$ column vector with zeros in all components that is used to suppress the undesired targets in $\mathbf{U}$.
Analogous to Feng et al. \cite{feng2015random}, Du et al. \cite{du2011random} conducted target detection by TCIMF, where RP is used for dimensionality reduction.
Not only does RP reduce the computational complexity, but it improves the target-detection accuracy by decision fusion across multiple RP instances.
The experimental results demonstrated that the detection accuracy of RP-TCIMF with a single run of RP is slightly lower than that of TCIMF, but it can be further improved by performing multiple runs.

Subspace-based spectrum estimation is often used for wideband spectrum sensing.
However, subspace-based techniques, which require eigendecomposition of the original data, are computationally expensive.
To alleviate this issue, Majee et al. \cite{majee2015efficient} applied low-rank approximation before multiple signal classification (MUSIC) \cite{schmidt1986multiple}, which is employed for wideband spectrum sensing.
The low-rank approximation technique used in this work is based on Cholesky factorization (CF) \cite{halko2011finding} and RP.
Suppose $\mathbf{X} \in \mathbb{R}^{n \times n}$ is an arbitrary square matrix.
First, RP is performed as $\widehat{\mathbf{X}} = \mathbf{X}\mathbf{W}$, where $\mathbf{W} \in \mathbb{R}^{n \times k}$ is a random projection matrix.
Then, $\mathbf{\Phi}^{\mathrm T}$ is filled with $k$ left singular values of $\widehat{\mathbf{X}}$; thus, $\widehat{\mathbf{X}}$ can be estimated as $\widehat{\mathbf{X}'} = \mathbf{\Phi}\mathbf{X}\mathbf{\Phi}^{\mathrm T}$.
Next, CF is applied as $\widehat{\mathbf{X}'} = \mathbf{L}\mathbf{L}^{\mathrm T}$, and $\mathbf{D}$ can be calculated as $\mathbf{D} = \widehat{\mathbf{X}'}\mathbf{\Phi}^{\mathrm T}(\mathbf{L}^{\mathrm T})^{-1}$.
SVD is then applied as $\mathbf{D} = \mathbf{U}\mathbf{\Sigma}\mathbf{V}^{\mathrm T}$, and the $k$-rank approximation of $\mathbf{X}$ can be obtained as $\mathbf{X}^{LR} = \mathbf{U}\mathbf{\Sigma}^2\mathbf{V}^{\mathrm T}$.
The authors concluded that the proposed method achieves a marginal reduction in time complexity.
In addition, the spectrum sensing performance of the proposed method is comparable or superior to that of MUSIC in terms of the probability of alarm.

\section{Dimensionality Increasing Approaches}

In some studies, researchers act in diametrically opposed ways.
They first map the original dataset onto a higher-dimensional feature space to better represent the original features.
Then, RP is applied to reduce the dimensionality and computational cost.

\begin{figure*}[!htb]
  \centering
  \resizebox{.7\linewidth}{!} {
    \includegraphics{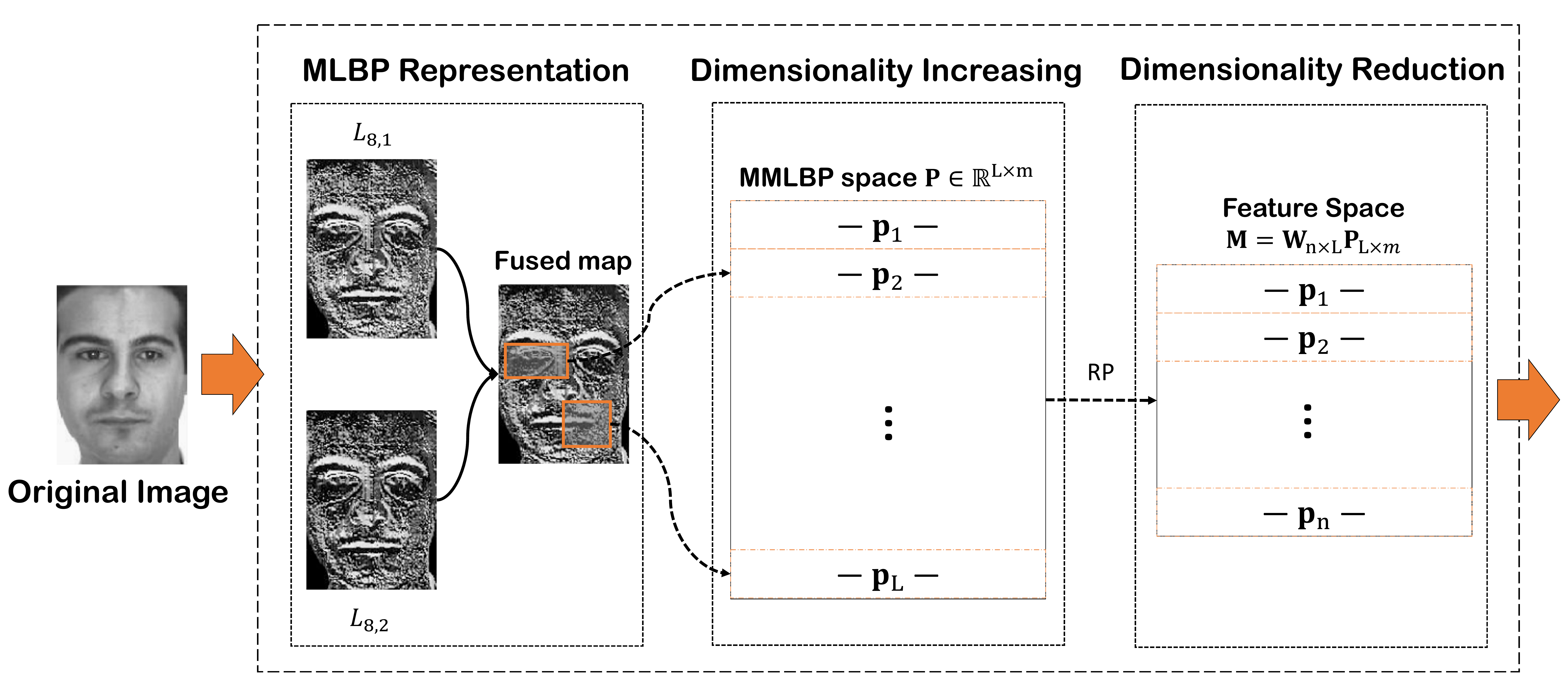}
  }
  \caption{Overview of a robust face recognition method. The facial features are mapped onto a very-high dimensional space by rectangle filters. Then, RP is used to compress features into a low-dimensional subspace.}
  \label{fig:robust-face-recognition-method-overview}
\end{figure*}

Ma et al. \cite{ma2015random} proposed a robust method for face recognition, as illustrated in Figure~\ref{fig:robust-face-recognition-method-overview}.
In the proposed method, multi-radius local binary pattern (MLBP) is first adopted to incorporate more structural facial features.
Then, a high-dimension multiscale and multi-radius LBP (MMLBP) space  $\mathbf{P} \in \mathbb{R}^{L \times m}$ is obtained by convolving rectangle filters.
The rectangle filters of a given face image with width $w$ and height $h$ can be defined as follows

\begin{equation}
  \mathbf{p}_i^{u_0, v_0} = 
  \begin{cases}
  1, & u_0 \le u \le w_i, v_0 \le v \le h_i\\
  0, & otherwise\\
  \end{cases}
  \label{eq:rectangle-filters}
\end{equation}
where $u_0$ and $v_0$ represent the offset coordinates of filter $p_i$ with width $w_i$ and height $h_i$.
There are approximately $L = m^2 = (wh)^2$ (i.e., the number of possible locations times the number of possible scales) exhaustive rectangle filters for each face image.
Generally, a large multiscale rectangle filter matrix $\mathbf{P}$ can be obtained by stacking $\mathbf{p}_i$ together, where $p_i \in \mathbb{R}^{1 \times m}$ can be formulated as a row vector whose dimensionality is equal to $w \times h$.
Since $L$ is usually $10^6-10^{10}$, sparse RP \cite{li2006very} is used to perform dimensionality reduction.
The subspace can be formulated as $\mathbf{M}_{n \times m} = \mathbf{W}_{n \times L}\mathbf{P}_{L \times m}$, where $\mathbf{W}$ is a transformation matrix generated by RP.
The proposed method not only achieves a higher recognition rate but also shows better robustness to corruption, occlusion, and disguise compared to Randomface \cite{wright2009robust} and Eigenfaces \cite{turk1991eigenfaces}, even in low-dimensional spaces.

Multiple studies have focused on visual tracking \cite{shan2016improved,zhang2012real,zhang2014fast}, where RP is favored by researchers because of its characteristics of computational effectiveness and data independence.
Zhang et al. \cite{zhang2012real,zhang2014fast} proposed a tracking framework, named CT tracker, where rectangle filters, following Eq. \ref{eq:rectangle-filters}, are adopted to form a very high-dimensional multiscale image feature vector and RP is applied to compress samples of foreground targets and the background.
The tracking window in the first frame is determined manually. To predict the target in the subsequent frame, positive samples are taken near the current target while negative samples are taken far from the current target.
Both positive and negative samples are used to update the Bayes classifier.
However, rectangle filters are sensitive to the presence of a few outliers in the samples.
To overcome this problem, a more stable model was proposed by Gao et al. \cite{gao2016extended}, where the rectangle filters are replaced by MSERs \cite{matas2004robust}.
To robustly adapt to variations in target appearance, a least squares support vector machine (LS-SVM) \cite{suykens2002weighted} is employed.
The authors stated that the proposed tracker outperforms the CT-based tracker \cite{zhang2014fast} in terms of mean distance precision and mean frame rate (FPS).
Nevertheless, MSER features have limited performances on blurred images \cite{sluzek2016improving}.
Therefore, the performance of the proposed tracking method may be seriously affected on blurred videos.

Recent studies have revealed that ELM performs well in regression and classification problems \cite{huang2006extreme}.
Additionally, compared to traditional algorithms, such as back-propagation (BP) and support vector machine (SVM), ELM provides not only good generalization performance but also fast learning speed.
Alshamiri et al. \cite{alshamiri2015combining} performed RP in conjunction with ELM for low-dimensional data classification. The method consists of two phases.
First, the original data $\mathbf{X} \in \mathbb{R}^{n \times d}$ are projected onto the subspace spanned by $L$ ($L \gg d$) using ELM.
The linear separability of the data is often increased by mapping the data onto a high-dimensional ELM feature space.
Then, RP is applied to reduce the dimensionality of the ELM feature space.
There is a slight improvement in the classification accuracy of ELM-RP compared to that of ELM.
Since ELM consists of a single hidden layer, it is difficult to encode complex things and achieve satisfactory accuracy.
Also, on small-n-large-p datasets, ELM is prone to overfitting because of the lack of training samples \cite{zhong2014comparing}.

\section{Ensemble Approaches}

Inspired by the fact that ensemble methods can significantly improve the performance of weak classifiers, several algorithms were proposed based on ensembles of decision trees. Well-known methods include random forest \cite{liaw2002classification} and AdaBoost \cite{freund1999short}.
Several studies have proved that ensembles of multiple RP instances help RP to produce more stable results.

Schclar et al. \cite{schclar2009random} proposed an ensemble method based on RP and nearest-neighbor (NN) inducers \cite{cover1967nearest}. 
First, $K$ random matrices are generated by RP.
Then, $K$ training sets are constructed for the ensemble classifiers by applying random matrices to the original dataset.
Next, $K$ NN classifiers are trained, and the final classification result is produced via a voting scheme.
The proposed method is more accurate than the non-ensemble NN classifier.

Zhang et al. \cite{zhang2017drugrpe} proposed an RP ensemble method that is analogous to the work of Schclar et al. \cite{schclar2009random} for drug-target interaction prediction, where the ``PaDEL-Descriptor'' \cite{yap2011padel} is adopted as a feature detector.
The authors stated that the proposed drug-target interaction prediction method improves the accuracy by 4.5\%-8.2\% compared to the work of \cite{he2010predicting}.
Similarly, Yoshioka et al. \cite{yoshioka2013evaluation} proposed an RP ensemble method for dysarthric speech recognition, in which an automatic speech recognition (ASR) system is adopted as a feature detector.
Compared to the PCA-based feature projection method, the method proposed in \cite{yoshioka2013evaluation} improves the recognition rate by 5.23\%.

Gondara \cite{gondara2015rpc} also proposed an ensemble classifier using RP.
In contrast to the method proposed by Schclar et al. \cite{schclar2009random}, where the RP matrices are applied to the same feature space, in this method, the RP matrices are applied to random subsets of the original feature set.
The authors demonstrated that the proposed method performs equally well or better than random forest and AdaBoost in terms of classification accuracy.

To alleviate the high distortion in a single run of a clustering algorithm in the feature spaces produced by RP, Fern et al. \cite{fern2003random} investigated how RP could best be used for clustering and proposed a cluster ensemble approach based on expectation-maximization (EM) \cite{dempster1977maximum} and RP.
In the proposed ensemble approach, EM generates a probabilistic model $\theta$ of a mixture of $k$ Gaussian distributions after each RP run.
The final clusters are aggregated by measuring the similarity of clusters in different clustering results, where the similarity of two clusters $c_i$ and $c_j$ can be defined as

\begin{equation}
    sim(c_i, c_j) = \min_{p_i \in c_i, p_j \in c_j} P_{ij}^{\theta}
\end{equation}
where $P_{ij}^{\theta}$ denotes the probability of data points $i$ and $j$ belonging to the same cluster, which can be calculated as

\begin{equation}
    P_{ij}^{\theta} = \sum_{l=1}^k P(l | i, \theta) P(l | j, \theta)
\end{equation}
The authors noted that the proposed ensemble method (RP + EM) is more robust and produced better clusters than those of PCA + EM.
However, EM algorithm often gets stuck in a local minimum, even on perfect datasets \cite{wang2006severity}.

\section{Conclusions and future perspectives}
RP is an efficient and powerful dimensionality reduction technique that has been developed and matured during the past 15 years.
With the rapid increase of big data, RP has provided tangible benefits to counteract the burdensome computational requirements and has met the needs of real-time processing in some situations.
Despite the fact that RP is computationally efficient, it often introduces relatively high distortion.
To solve this problem, major successful efforts have been made to improve the performance of RP, as summarized in this survey.

The features cannot be well-represented in the high-dimensional space produced by extant methods, such as rectangle filters and ELM.
Therefore, the proposal of accurate dimensionality increasing approaches constitutes a promising direction for future RP research.

Other opportunities for future RP research will be the extension towards upcoming tasks that require real-time computation, such as  speech, voice, image and video recognition.
In these tasks, we believe RP will play an important role in handling the high-dimensional characteristics of these applications.

\section*{Acknowledgment}
This work is partially supported by the National Natural Science Foundation of China (Grant Nos. 61471147, 61371179), the Natural Science Foundation of Heilongjiang Province (Grant No. F2016016), the Fundamental Research Funds for the Central Universities (Grant No. HIT.NSRIF.2017037), and the National Key Research and Development Program of China (Grant No. 2016YFC0901905).

\ifCLASSOPTIONcaptionsoff
  \newpage
\fi



%
\bibliographystyle{IEEEtran}
\bibliography{references}

\end{document}